\documentclass{article}
\usepackage{graphicx}
\usepackage{float}
\usepackage{wrapfig}
\usepackage{subcaption}

\PassOptionsToPackage{numbers,sort&compress}{natbib}

\usepackage[preprint]{neurips_2026}
\makeatletter
\renewcommand{\@noticestring}{}
\makeatother

\usepackage[utf8]{inputenc} 
\usepackage[T1]{fontenc}    
\usepackage{hyperref}       
\usepackage{url}            
\usepackage{booktabs}       
\usepackage{amsfonts}       
\usepackage{nicefrac}       
\usepackage{microtype}      
\usepackage{xcolor}         

\usepackage{amsmath, amssymb}

\title{Learn to Match: Two-Sided Matching with Temporally Extended Feedback}

\author{
  Haijing Zong$^{*1}$ \quad
  Yancheng Liang$^{*2}$ \quad
  Boyang Zhou$^{2}$ \quad
  Natasha Jaques$^{2}$ \\[0.5em]
  $^1$Department of Economics, University of Washington, Seattle, United States \\
  $^2$Paul G. Allen School of Computer Science \& Engineering,  \\ University of Washington, Seattle, United States \\
  $^*$Equal contribution. Correspondence: \texttt{zhaijing@uw.edu}
}

\begin{document}

\maketitle

\begin{abstract}
Two-sided matching markets often involve information that unfolds over time through interviews, repeated interaction, learning, and separation. Existing matching models typically reduce this process to immediate sub-Gaussian feedback about fixed preferences, missing settings where payoff-relevant information is revealed gradually and changes future matching decisions. We introduce a framework with \textbf{temporally extended feedback}, that formulates two-sided matching as a partially observable Markov game with costly pre-match screening, noisy post-match observations, evolving latent profiles, and endogenous continuation or dissolution. We instantiate this framework in \textsc{Learn2Match}, a multi-agent reinforcement-learning benchmark for dynamic matching markets. \textsc{Learn2Match} supports decentralized decision making over whom to interview, whom to match with, and when to dissolve a match, while evaluating policies using regret, social welfare, and an information-friction loss that measures the welfare gap caused by incomplete revelation of latent preferences. We find that independent PPO achieves higher cumulative social welfare and lower cumulative regret than the bandit-style CA-ETC baseline under temporally extended feedback, demonstrating the promise of MARL for dynamic matching markets. However, PPO still incurs higher information-friction loss, revealing that end-to-end MARL does not yet provide the coordinated exploration structure of matching-bandit methods. These results position \textsc{Learn2Match} as a benchmark for developing the next generation of matching-market algorithms: methods that are adaptive like RL agents, statistically disciplined like bandit algorithms, and structurally aware like stable-matching mechanisms. Please refer to \url{https://sites.google.com/view/learn-to-match/home} for the official website and the code link.
\end{abstract}

\section{Introduction}
\begin{figure}
        \centering
        \includegraphics[width=1\linewidth]{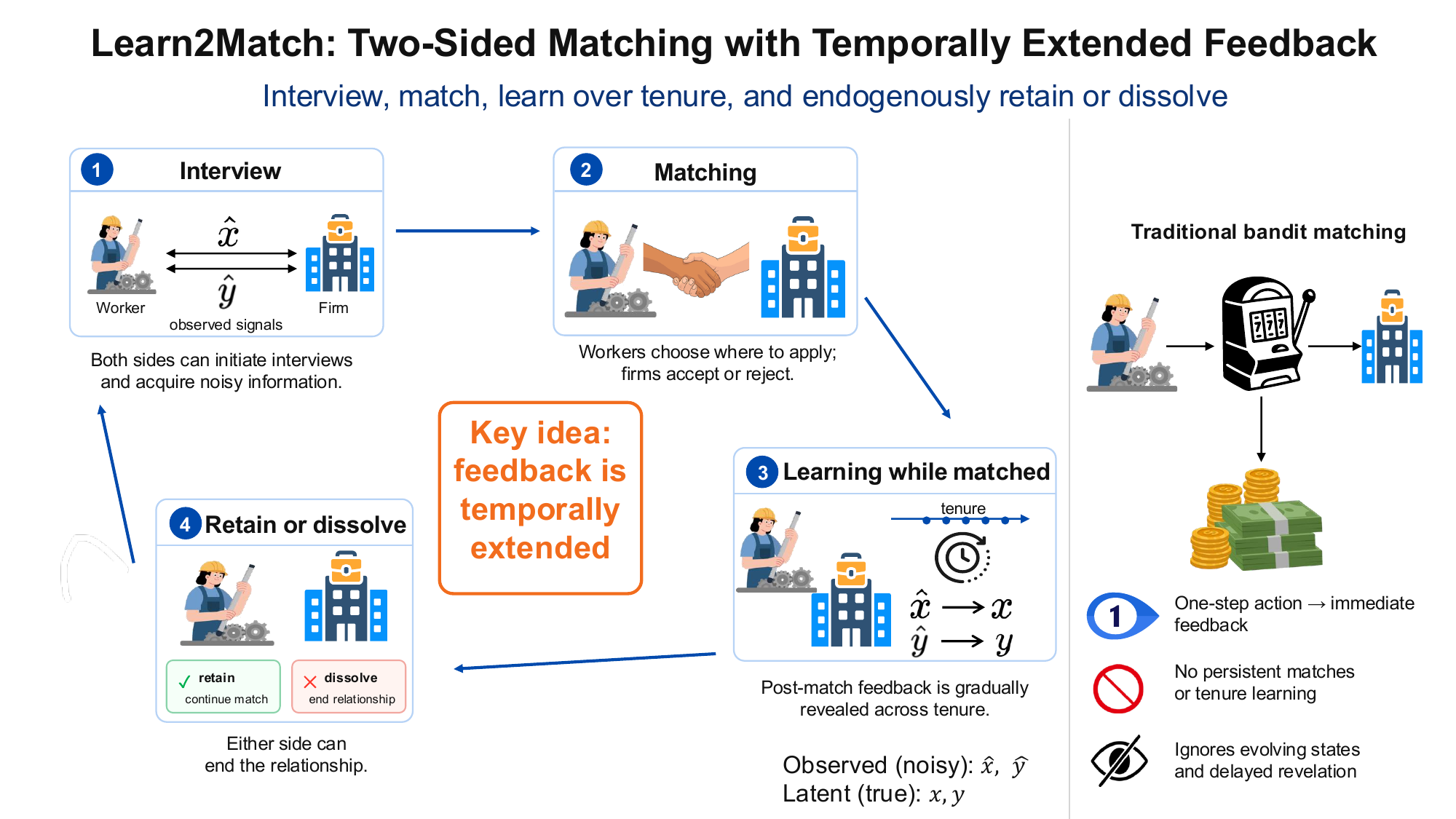}
        \caption{Overview of \textsc{Learn2Match}, a dynamic two-sided matching framework with temporally extended feedback. Agents interview, match, learn gradually during tenure, and decide whether to retain or dissolve relationships, in contrast to traditional bandit matching with immediate one-step feedback.}
\label{fig:learn2match_overview}
\end{figure}

Two-sided matching studies how agents on opposite sides of a market form pairwise relationships based on preferences over one another.
Its central solution concept is stability~\citep{gale1962college}: no unmatched pair should mutually prefer each other to their assigned partners.
The classical Gale--Shapley framework and deferred-acceptance algorithm form the foundation for this line of work, with applications in labor markets, residency matching, school choice, marriage and dating market~\citep{gale1962college,roth1992two,roth1996national,abdulkadirouglu2003school, hitsch2010matching, tu2014online}.

However, many real-world matching markets depart from the static Gale--Shapley model~\citep{abdulkadirouglu2003school,immorlica2021designing,azagirre2024better,yan2019interview,shi2025optimal}.
Recent work has therefore extended stable matching to incorporate learning, decentralization, uncertainty, and contextual information~\cite{jagadeesan2021learning,liu2021bandit,basu2021beyond,kong2023player,pokharel2023converging,pagare2023two,zhang2024decentralized,parikh2024competing,athanasopoulos2025probably,kong2025bandit,hosseini2024putting,li2022dynamic}.
These models narrow the gap between classical matching theory and operational matching systems, but they typically assume that each matching decision generates an immediate reward, observation, or noisy signal.

In many markets, however, the payoff-relevant information generated by a matching decision is realized only through future states.
We call this phenomenon \emph{temporally extended feedback}.
For example, in labor markets, match quality may evolve after employment begins as workers gain experience, firms learn about productivity, and both sides decide whether to continue or separate~\cite{farber1996learning,altonji2001employer,lange2007speed,schonberg2007testing,topel1992job,munasinghe2000wage}.
Similarly, pre-match processes such as search, screening, interviewing, and negotiation may consume substantial calendar time and effort before any match is formed~\cite{mccall1970economics,diamond1982aggregate,mortensen1994job,pissarides2000equilibrium,carrillo2023recruitment}.
Thus, matching decisions can affect future opportunities, information, and welfare through persistent state changes rather than through one-period feedback alone. In labor markets, such persistence is often described as \emph{career path dependence}: a match can shape future outcomes even without frequent rematching, as employment relationships become more durable with tenure~\citep{farber1996learning,topel1992job}. This path dependence arises because workers learn about fit and accumulate occupation-specific human capital over time, so current matches affect future productivity, information, and mobility costs~\citep{miller1984job,kambourov2009occupational,groes2015u}.

We propose to model two-sided matching markets with temporally extended feedback explicitly. In contrast to matching-bandit formulations, where each matching decision yields one-shot feedback about fixed underlying preferences, we cast the market as a partially observable Markov game: a sequential decision problem in which a latent state governs how observations, rewards, and matching opportunities evolve over time. This formulation captures three sources of temporal feedback: evolving agent profiles, costly and persistent pre-match or post-match stages, and noisy partial observations of latent match quality. Agents must therefore decide not only whom to match with, but also when to search, screen, continue, separate, and explore.

We instantiate this framework through \textsc{Learn2Match}, a benchmark for dynamic two-sided matching markets.
\textsc{Learn2Match} models agents with evolving latent profiles, costly pre-match and post-match decisions, noisy observations, and endogenous continuation or separation.
We evaluate multi-agent reinforcement-learning methods~\cite{lowe2017multi,rashid2020monotonic,zhang2021multiagent,yu2022surprising,schulman2017proximal} against bandit-style baselines adapted from the matching literature~\cite{pagare2024explore,mirfakhar2026bandit}.
Our evaluation includes matching regret, social welfare, and an \emph{information-friction loss}: the gap between acting on current observed or believed preferences and the stable matching induced by the true latent preferences revealed through sufficient interaction.

We test independent PPO, a state-of-the-art multi-agent reinforcement learning (MARL) algorithm~\citep{schulman2017proximal, yu2022surprising}, on \textsc{Learn2Match}, and compare it with a bandit-style baseline CA-ETC~\citep{pagare2024explore}. We show that MARL agents perform better in cumulative social welfare and reduce the cumulative regret, showing the promise of MARL for decision-making in matching markets under temporally extended feedback. Notably, despite having higher performance, the information-friction loss of MARL remains higher compared to the bandit-style algorithm, which is designed to have a coordinated exploration process by its algorithm.
More broadly, \textsc{Learn2Match} opens a shared research agenda between MARL and matching theory: it challenges MARL community to build agents that coordinate exploration and exploitation in structured markets, while challenging the matching and bandit communities to move beyond one-shot preference estimation toward algorithms that remain effective when information unfolds through time.
In this sense, \textsc{Learn2Match} is not only a benchmark for evaluating existing methods, but also a call to design the next generation of learning algorithms for matching markets: algorithms that are as adaptive as RL agents, as statistically disciplined as bandit methods, and as structurally aware as stable-matching mechanisms.

Our contributions are threefold.
First, we introduce \textsc{Learn2Match}, a benchmark for dynamic two-sided matching with temporally extended feedback, featuring multi-round information acquisition, evolving agent profiles, and endogenous proposal, response, and dissolution decisions.
Second, we provide an initial empirical study comparing multi-agent reinforcement learning with bandit-based baselines. Third, \textsc{Learn2Match} opens a shared research agenda at the intersection of multi-agent reinforcement learning and matching theory, and flexibly supports real-world application modeling---from labor markets and residency matching to school choice, dating, and ride-hailing dispatch. Progress on this benchmark can translate into better outcomes wherever the static, fully revealed assumptions of classical matching theory fail to hold.

\section{Related work}

\textbf{Stable matching with static information.}
Two-sided matching has classically been studied through the lens of stability, beginning with the Gale--Shapley framework and its applications to labor markets, school choice, and other centralized assignment settings~\citep{gale1962college,roth1992two,roth1996national,abdulkadirouglu2003school}. Existing work studies how stable matchings or preferences can be learned from noisy observations, but typically assumes an underlying static match value or preference structure.~\citep{jagadeesan2021learning,liu2021bandit,basu2021beyond,kong2023player,pagare2023two,pokharel2023converging,zhang2024decentralized,parikh2024competing,kong2025bandit,athanasopoulos2025probably,li2022dynamic,hosseini2024putting}. These models typically reduce uncertainty to noisy (like zero-mean sub-Gaussian) observations of fixed latent preferences, with feedback revealed immediately after each interaction. This is insufficient for settings where match quality evolves through interaction and history. For example, in a dating market, partners' profiles evolve as they age and accumulate life experience. Our work builds on this literature while shifting attention from stable-outcome identification alone to a richer sequential environment in which agents choose how to acquire information, whether to match, and whether to remain matched over time.

\textbf{Frictional search and post-match learning} are widely studied by a large economics literature~\citep{mccall1970economics,diamond1982aggregate,mortensen1994job,pissarides2000equilibrium,shimer2005cyclical,carrillo2023recruitment}, which emphasizes that employment matching are not instantaneous assignment. These markets are shaped by costly vacancy posting, screening, and hiring frictions~\citep{barron1997employer,blatter2012costs,huang2006employee}, and are followed by employer learning, human-capital accumulation, turnover, and costly post-match adjustment~\citep{mincer1974schooling,farber1996learning,altonji2001employer,lange2007speed,schonberg2007testing,topel1992job,munasinghe2000wage,silva2009labor}. Recent works on stable matching~\citep{ashlagi2025stable,allman2025signaling,mirfakhar2026bandit,babaioff2026efficient} treat screening and in,formation as complementary pre-match signals to reveal the underlying preference and fails to capture that information can arrive after the match initiates. \textsc{Learn2Match} fills that gap and turns the above mechanisms into a controlled multi-agent learning environment. These models, however, are stylized analytical tools that isolate single mechanisms under tractability assumptions to explain aggregate behavior; they are not controlled environments in which agents \emph{learn} matching strategies from interaction. \textsc{Learn2Match} fills that gap: a simulatable benchmark grounded in this literature. Although labor markets are our leading example, the same structural features---persistent state, costly information acquisition, and gradual revelation of match quality---arise in dating, school choice, and platform-based matching, where the framework applies equally

\textbf{Dynamic matching and multi-agent reinforcement learning (MARL).}
\citet{min2022learn} initiated a theoretical study of RL on matching markets. While their model allows the market context to evolve across rounds, each round still implements a centralized myopic stable matching planner and receives immediate noisy utility feedback for the matched pairs. Our worker instead adopts a fully decentralized decision-making paradigm with temporally extended feedback. \citep{liu2022welfare} study welfare maximization in Markov exchange economies, and \citep{shi2023multiagent} develop MARL frameworks for off-policy evaluation in two-sided markets. More broadly, modern MARL provides tools for decentralized partially observable multi-agent decision-making~\citep {lowe2017multi,rashid2020monotonic,zhang2021multiagent,yu2022surprising,schulman2017proximal}, and has been used to simulate market-scale systems such as tax policy and mechanism design~\citep{zheng2020ai,karten2025llm}.
Our contribution differs in three concrete axes. First, we treat the market as a partially observable Markov \emph{game}---\textbf{decentralized two-sided learning} rather than a centralized planner with fully observable states. Second, we model the specific mechanisms that produce temporally extended feedback in real matching markets: multi-round pre-match screening, gradual post-match revelation through tenure, and endogenous dissolution, none of which appear together in the works above. 

\section{Preliminary}

We study a two-sided dynamic matching market with strategic agents on both sides or on one side only. There are $N^{\mathcal W}$ workers on side $\mathcal W$ and $N^{\mathcal F}$ firms on side $\mathcal F$ (which can also represent, e.g., students and schools, patients and hospitals, or women and men). Agents on each side may propose to, accept, reject, or dissolve matches with agents on the opposite side.

We denote $[N]=\{1,2,...,N\}$. For a set $S$, $\Delta(S)$ is the set of all probability distributions over $S$. If not stated otherwise, agent $i$ without any specification refers to an agent either on the $\mathcal W$ or $\mathcal F$ side, for notation simplicity.

To capture such delayed and sequentially revealed interaction feedback, we model the matching market as a partially observable Markov game. Compared to repeated tabular games or bandit algorithms, a key property of Markov games is the capability to model sequential dynamics, which is necessary for a matching market with delayed feedback. We formulate the environment as:

\[
\mathcal G = \left( N^{\mathcal W}, N^{\mathcal F}, \mathcal S, \mathcal P, \{\mathcal O_i^{\mathcal W}, \mathcal A_i^{\mathcal W}, R_i^{\mathcal W}\}_{i=1}^{N^\mathcal W}, \{\mathcal O_j^{\mathcal F}, \mathcal A_j^{\mathcal F}, R_j^{\mathcal F}\}_{j=1}^{N^\mathcal F}, \rho_0 \right).
\]

Here $\mathcal S$ is the state space, and $\mathcal A^{\mathcal W}=\mathcal A^{\mathcal W}_1\times...\times\mathcal A^{\mathcal W}_{N^\mathcal{W}}$ and $\mathcal A^{\mathcal F}=\mathcal A^{\mathcal F}_1\times...\times\mathcal A^{\mathcal F}_{N^\mathcal{F}}$ denote the joint action spaces of workers and firms, respectively. On state $s$, each agent $i$ receives the observation $o_i=\mathcal O_i(s)\in O_i$ by the observation mapping function $\mathcal O_i$ and takes an action $a_i$ following the policy $\pi_i:O_i\to \Delta(\mathcal A_i)$. With the joint action $\mathbf a^{\mathcal W}=\left(a^{\mathcal W}_1, ..., a^{\mathcal W}_{N^{\mathcal W}}\right)$ and $\mathbf a^{\mathcal F} = \left(a^{\mathcal F}_1, ..., a^{\mathcal F}_{N^{\mathcal F}}\right)$, the next state $s' \sim \mathcal P(s, \mathbf a^{\mathcal W}, \mathbf a^{\mathcal F})$ follows the transition function $\mathcal P:S\times \mathcal A^{\mathcal W}\times \mathcal A^{\mathcal F} \to \Delta(\mathcal S)$. And each agent $i$ receives a reward $r_i$ given by the reward function $R_i:\mathcal S\times \mathcal A^{\mathcal W}\times \mathcal A^{\mathcal F} \to \mathbb R$.

\section{Problem setting}
\label{sec:problem-setting}
Bandit-based matching formulations assume immediate feedback, in which the consequences of proposals, acceptances, and separations are realized within the same round. Instead, our formulation allows delayed feedback throughout the matching process. We instead thread partial observability through every stage; An offer need not be resolved immediately but may enter an intermediate evaluation stage whose outcome is revealed only after several periods; And even after a match is formed, match quality may remain partially observed and continue to evolve over time. Continuation and dissolution decisions therefore depend on information that arrives gradually rather than immediately.

\textbf{State.}
At each step, the market is in a state
\[
    s = (X(s), Y(s), M(s), H(s)))
\]
Here, $X(s)$ and $Y(s)$ represent the information (which can be latent, believed, or observed) of the agents of side $\mathcal W$ or $\mathcal F$, respectively. To be specific, $x_i,y_j\in \mathbb R^d$ are the latent profiles of agent $i$ on side $\mathcal W$ and agent $j$ on side $\mathcal F$, respectively. On state $s$, agent $j$ on side $\mathcal F$ has an observation or belief of agent $i$ on side $\mathcal W$, which is denoted as $\hat x_{ij}\in\mathbb R^d$. $\hat y_{ij}$ is defined similarly.

The matching set
\[
    M(s) \in [N^{\mathcal W}] \times [N^{\mathcal F}]
\]
denotes all matched pairs $(i,j)$ for the current state $s$.

The history term $H(s)$ summarizes the relevant past interactions in the market, such as previous matches, past rewards, interview outcomes, proposal and rejection decisions, and other observations generated by the environment. We do not give a strict form of $H(s)$ here to allow for flexible control of the environment dynamics.

\textit{\textbf{Asymptotic Revelation.}}
Motivated by economics literature \citep{cripps2008common, farber1996learning, lange2007speed}, we assume that if a pair $(i,j)$ is matched for infinite times, the latent variable will be revealed for both agent (up to a constant $\epsilon)$.
\begin{align}
    \label{def:reveal}
    & \forall i\in [N^{\mathcal W}],j\in [N^{\mathcal F}], (s_1,s_2,...)\in \mathcal S^\infty, \notag \\
    & \left( \sum_{t=0}^T [(i,j)\in M(s_t)] \right) =\infty \Rightarrow \lim_{t\to\infty} |\hat x_{i,j}(s_t)-x_i| \le \epsilon ~\text{and}~ \lim_{t\to\infty} | \hat y_{i,j}(s_t) - y_j | \le \epsilon. \notag
\end{align}

\textbf{Action.}
We allow three broad categories of actions: \emph{proposal}, \emph{response}, and \emph{dissolution}.

\textit{Proposal.} An agent may initiate an interaction with an agent on the opposite side. There could be multiple pre-matching screening stages. For example, in the labor market, the interview is the main screening stage. In school application, there can be multiple screening stages, like application review and interview.

\textit{Response.} Upon receiving a proposal, the receiving agent chooses whether to \emph{accept} or \emph{reject} it. If a screening invitation is accepted, the two agents enter the screening stage; if a match proposal is accepted by both sides under the market rules (optional environment design choice, e.g., match can only be accepted after interview), the pair becomes matched and will be added to $M(s)$. 

\textit{Dissolution.} Either side may dissolve an existing match at any time, potentially incurring a cost. This captures quitting, firing, churn, or other forms of separation.

In this work, we assume only agents on side $\mathcal W$ can propose a matching. (Such constraints do not apply to screening proposals.) This assumption follows the population agent--arm formulation commonly adopted in bandit learning for two-sided matching markets
\citep{liu2021bandit, basu2021beyond, li2022dynamic, kong2023player, pagare2024explore, li2025survey, mirfakhar2026bandit},
and enforces the one-to-one matching constraint at the environment level by preventing duplicate assignments, i.e., matching one agent to multiple agents on the other side in the same round.

\textit{\textbf{Step or Time Unit.}} In this work the time step $t$ can be regarded as the minimal time unit or division (e.g., a calendar month) for matching. That is, no changes of matching can happen inside the same $t$. Except from that, we note that within the same $t$ there can be multiple actions taken ``simultaneously''. For example, a pair of agents can initiate a screening and accept it in the same step. We consider a finite horizon $T$ for each episode.

\textbf{Reward.} We consider a linear reward model with unobservable latent profiles of agents and observable empirical variables of agents. To be specific, for a matched pair $(i,j)$, agent $i$ on side $\mathcal W$ receives
$R^{\mathcal W}_i(s)=\langle x_i,\hat y_{ij}(s)\rangle$
and agent $j$ on side $\mathcal F$ receives a reward $R^{\mathcal F}_j(s) = \langle \hat x_{ij}(s), y_j \rangle$ on state $s$. If an agent is not matched on the current step $t$, it receives zero reward. We remark that linear rewards are widely used in real-world matching markets such as recommendation and advertising \citep{li2021tight, guo2021we}, and are general enough to represent complete ordinal preference rankings, i.e., strict preference lists, as used in the classic Gale--Shapley stable matching model \citep{gale1962college}.


For MARL, the objective (value function) of $\pi_i$ is
\[
V_i^{\pi_i,\pi_{-i}} = \mathbb E \left[ \sum_{t=1}^T R(s_t,\mathbf{a}_t) \mid (s_1,\mathrm{a}_1,...,s_T,\mathrm{a}_T) \sim \pi_i,\pi_{-i} \right].
\]

Therefore, the value of $\pi_i$ depends not on the policy itself but also $\pi_{-i}$, the policies of all other agents on both sides, as is typical of multi-agent Markov games.

\textbf{Evaluation metric.}
Following prior work on bandit learning in decentralized matching markets
\citep{liu2021bandit, basu2021beyond, kong2023player, pagare2024explore},
we adopt the worker-optimal regret and firm-pessimal regret.
Let $M^*$ be the $\mathcal W$-optimal stable matching, equivalently the worker-optimal
and firm-pessimal stable matching, as characterized by the Gale--Shapley framework
and the lattice structure of stable matchings \citep{gale1962college, roth1992two}. The regrets are
\begin{align*}
	\overline{\textit{Regret}^{\mathcal W}} =& \sum_{t=1}^T  \left( \sum_{(i,j)\in M^*}^{N^{\mathcal W}} \langle x_i,y_{j} \rangle - \sum_{(i,j)\in M(s_t)} \langle x_i,\hat y_{ij}(s_t) \rangle \right), \\
	\underline{\textit{Regret}^{\mathcal F}} =& \sum_{t=1}^T  \left( \sum_{(i,j)\in M^*} \langle x_i,y_{j} \rangle - \sum_{(i,j)\in M(s_t)} \langle \hat x_{ij}(s_t), y_j \rangle \right).
\end{align*} 

We also consider \textit{Social Welfare} as the total reward of all agents.
\[
	\textit{Social Welfare} = \sum_{t=1}^T \sum_{(i,j)\in M(s_t)} \langle x_i, \hat y_{ij}(s_t) \rangle + \langle \hat x_{ij}(s_t), y_j \rangle.
\]

To capture temporally extended feedback, we consider a specific \textit{Friction Loss} as the loss of social walfare due to incorrect belief, assessment or matching decisions. Let $M^*(s_t)$ be the $\mathcal W$ worker-optimal stable matching under the preference defined by $\langle x_i,\hat y_{ij}(s_t)\rangle ,\langle \hat x_{ij}(s_t), y_j \rangle$.
\[
	\textit{Fricition Loss} = \sum_{t=1}^T \left( \sum_{(i,j)\in M^*} \langle x_i,y_j \rangle - \sum_{(i,j)\in M^*(s_t)} \langle x_i,y_j \rangle  \right).
\]

The \textit{Friction Loss} captures how the gap  on delayed observation can cause a gap on the final outcome (reward or social walfare).

\section{\textsc{Learn2Match}: A MARL Benchmark for Dynamic Two-Sided Matching}
\label{sec:Learn2Match}

We instantiate the framework in Section~\ref{sec:problem-setting} as \textsc{Learn2Match}, a multi-agent reinforcement-learning benchmark for two-sided matching with temporally extended feedback. The environment models a market in which agents screen potential partners, form matches, learn gradually from interaction, and may endogenously dissolve matches. 

\subsection{Order of action and reward settlement}
Each step $t$ in \textsc{Learn2Match} consists of five phases. First, unmatched agents on both sides simultaneously propose interviews. Second, each agent accepts at most one received interview proposal; accepted pairs $(i,j)$ generate noisy observations but no match or production reward, so screening only refines beliefs~\citep{mirfakhar2026bandit, ashlagi2025stable, allman2025signaling, babaioff2026efficient}. Third, unmatched workers may propose matches only to previously interviewed firms. Fourth, an unmatched firm accepts or rejects the match proposal; accepted pairs are added to the current matching. Finally, either party in a matched pair may dissolve the match. Rewards are earned only by pairs that remain matched through step $t$.

\subsection{Life cycle of a matching}

At the beginning of an episode, the latent profiles $ x_i$ and $ y_j$ are sampled i.i.d. from $ N(0,I_d)$ for each agent. In this work, we consider $d=3$, but people can feel free to try different parameters. 

We will show the dynamics of the environment from the perspective of a pair $(i,j)$ going through the entire process of interview, matching, and dissolution. For any state $s$, let $\text{tenure}_{ij}(H(s))$ be the number of steps that $(i,j)$ are matched until state $s$.

Suppose at step $t_1$, $(i,j)$ is not matched. (That is, $(i,j)\notin M(s_t)$.) Then $i$ can initialize an interview proposal to $j$. If $j$ accepts the interview, both agents observes 
\[
    \hat{x}_{ij}(s_{t_1})=x_i+\varepsilon^{\mathcal W}_{\text{interview},ij}(s_{t_1}),
    \qquad
    \hat{y}_{ij}(s_{t_1})=y_j+\varepsilon^{\mathcal F}_{\text{interview},ij}(s_{t_1}),
\]
where all noise terms $\varepsilon_{\text{interview}}$ are i.i.d. sampled from $\mathcal N(0,\sigma_\text{interview}^2)$. As the number of interviews increases, the accumulated noisy observations become increasingly concentrated around the underlying latent profiles, enabling agents to infer the latent profiles more accurately from the interaction history.

Based on the observation history through the interview, the agents can decide to form a match, proposed by agents on side $\mathcal W$ and responded by agents on side $\mathcal F$. Suppose on time $t_2 \ge t_1$, $(i,j) \in M(s_{t_2})$, both agents observe
\[
\begin{aligned}
\hat x_{ij}(s_{t_2})
&= \frac{x_i}{1 + e^{-\lambda_\mathrm{reveal} \cdot \text{tenure}_{ij}(H(s_{t_2}))}}
+ \epsilon^{\mathcal W}_{\text{match},ij}(s_{t_2}), \\
\hat y_{ij}(s_{t_2})
&= \frac{y_j}{1 + e^{-\lambda_\mathrm{reveal} \cdot \text{tenure}_{ij}(H(s_{t_2}))}}
+ \epsilon^{\mathcal F}_{\text{match},ij}(s_{t_2}), ~~~\sigma_\text{interview}^2
> \sigma_\text{match}^2,
\end{aligned}
\]
and receives rewards according to the above observation. Such dynamics models how the agent's profile, belief, or observation may vary over time and how that converges to the latent profile with longer interaction. Therefore, a match may be locally acceptable under current beliefs but later become undesirable as new information arrives or states evolve. Thus, post-match learning and dissolution cannot be reduced to a one-shot stable-matching outcome, consistent with employer-learning and turnover models~\citep{farber1996learning, altonji2001employer, lange2007speed, topel1992job}. The parameters $\lambda_\mathrm{reveal}$ control how quickly realized states approach latent capacities while matched, capturing learning-by-doing, human-capital accumulation, firm growth, or related productivity changes~\citep{mincer1974schooling, farber1996learning, altonji2001employer, lange2007speed, schonberg2007testing}.

The pair $(i,j)$ can dissolve at some future step $t_3\ge t_2$. Each matched agent observes its current partner, tenure, latest estimate, and recent reward, then chooses whether to keep or dissolve the match. Dissolution occurs if either side chooses to dissolve. A dissolved agent becomes unmatched before entering a new match; stricter constraints, such as requiring exit before new interviews, can also be imposed.

Rewards are computed only after the dissolution phase. (In the example above, rewarding period is $t_2,t_2+1,...,t_3-1$. Hence, a match that forms and dissolves within the same time step generates no reward, imposing an opportunity cost on failed match formation.

\section{Experiments}
\label{sec:experiments}

\subsection{Algorithms}

We use \textsc{Learn2Match} to study how temporally extended feedback affects different decision-making algorithms. We compare one classic bandit-style stable matching algorithm with an end-to-end MARL algorithm.

\textbf{Independent PPO.}
We use the state-of-the-art MARL algorithm independent proximal policy optimization (PPO)~\citep{schulman2017proximal,yu2022surprising} and adopt a decentralized implementation. Each agent observes its local state---current match status, available proposals, interview history, belief estimates, tenure, and recent rewards---and outputs actions over interview proposals, match proposals or responses, and dissolution. Agents are trained to maximize cumulative reward in an episode.

\textbf{CA-ETC~\citep{pagare2024explore}.}
We adapt CA-ETC~\citep{pagare2024explore} to \textsc{Learn2Match} by treating workers as players and firms as arms, executed entirely through the environment's native action protocol. In the exploration stage of CA-ETC, a collision-avoiding round-robin schedule routes every worker--firm pair through the legal interview $\rightarrow$ match-proposal $\rightarrow$ acceptance sequence so that empirical estimates of $R_i^{\mathcal W}$ and $R_j^{\mathcal F}$ are recorded once at match formation. In the exploitation stage of CA-ETC, agents construct preference lists from estimated reward and execute the worker-proposing Gale--Shapley by issuing the corresponding proposals and acceptances. Subsequent dissolution is governed by the \textsc{Learn2Match} transition dynamics and does not affect CA-ETC's preference estimation; the adaptation thus preserves the original explore-then-commit structure while respecting the benchmark's interaction constraints. Implementation details are in the Appendix.

\subsection{Environment Setups}

In the experiments, small market uses $N^{\mathcal W}=N^{\mathcal F}=5$, $d=3$, $T=200$ and large market uses $N^{\mathcal W}=N^{\mathcal F}=20$, $d=5$, $T=600$. 
We consider the following two settings:

\textbf{\textit{Low-noise / near-static.}}
    A near-noiseless configuration with $\sigma_{\text{interview}}=10^{-6}$, $\sigma_{\text{match}}=0$, and $\lambda_{\text{reveal}}=100$. In the notation of Section~\ref{sec:Learn2Match}, the large $\lambda_{\text{reveal}}$ makes the sigmoid factor $1/(1+e^{-\lambda_\mathrm{reveal} \cdot \text{tenure}})$ saturate to one essentially instantaneously, so latent profiles are recoverable from a single interview and post-match observations converge to the latent profiles immediately. This setting matches the information structure assumed by decentralized matching-bandit algorithms and is intentionally favorable to CA-ETC. 
    
  \textbf{\textit{Temporally extended feedback.}} 
    The full benchmark with $\sigma_{\text{interview}}=1$, $\sigma_{\text{match}}=0.6$, and $\lambda_{\text{reveal}}=1$. We run such a setting with both small and large markets. Thus, $\hat x_{ij}$ and $\hat y_{ij}$ approach their latent values only after several periods of tenure rather than instantaneously. The latent stable matching is no longer recoverable from a single round of interviews.

\subsection{Evaluation metrics}

We evaluate both PPO agents and CA-ETC by worker-side regret, firm-side regret, social welfare, and friction loss as defined in Section~\ref{sec:problem-setting}, and provide training curves for PPO.

\subsection{Results}
\begin{figure}[htbp]
    \centering
    \begin{subfigure}{0.24\linewidth}
        \centering
        \includegraphics[width=\linewidth]{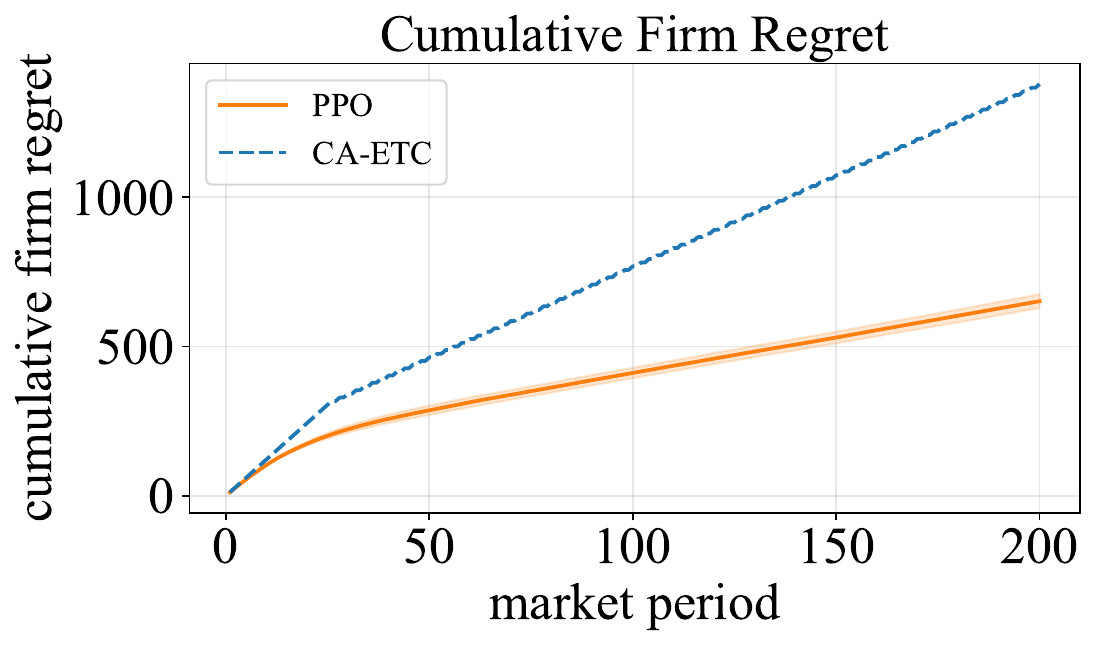}
        \caption{Firm regret}
        \label{fig:firm_regretno_noise}
    \end{subfigure}
    \hfill
    \begin{subfigure}{0.24\linewidth}
        \centering
        \includegraphics[width=\linewidth]{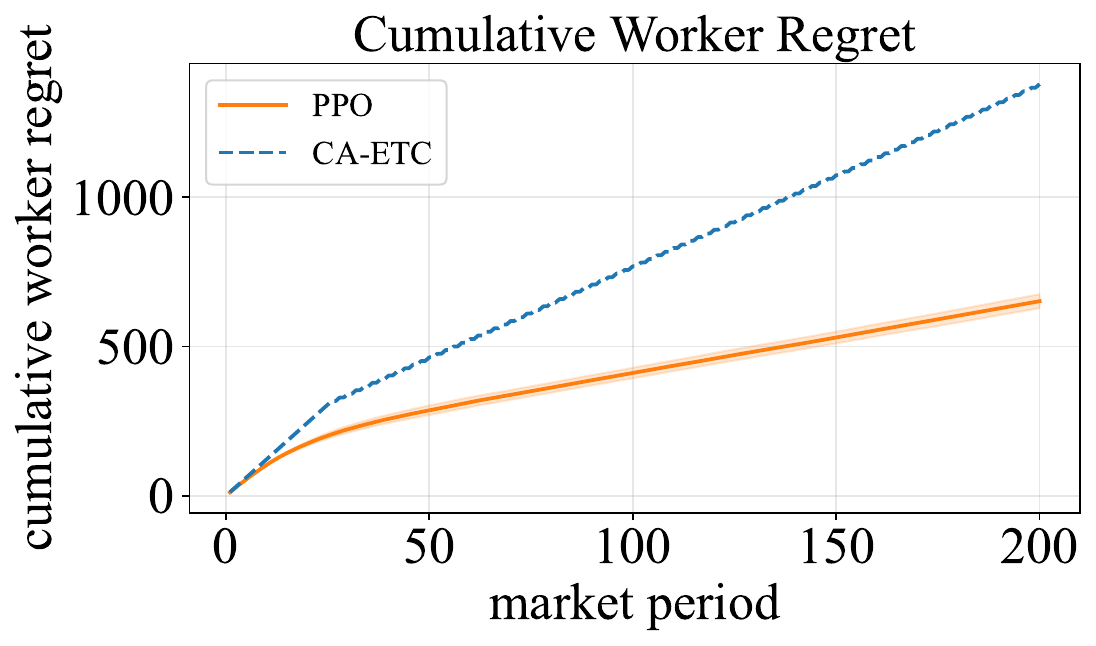}
        \caption{Worker regret}
        \label{fig:worker_regretno_noise}
    \end{subfigure}
    \hfill
    \begin{subfigure}{0.24\linewidth}
        \centering
        \includegraphics[width=\linewidth]{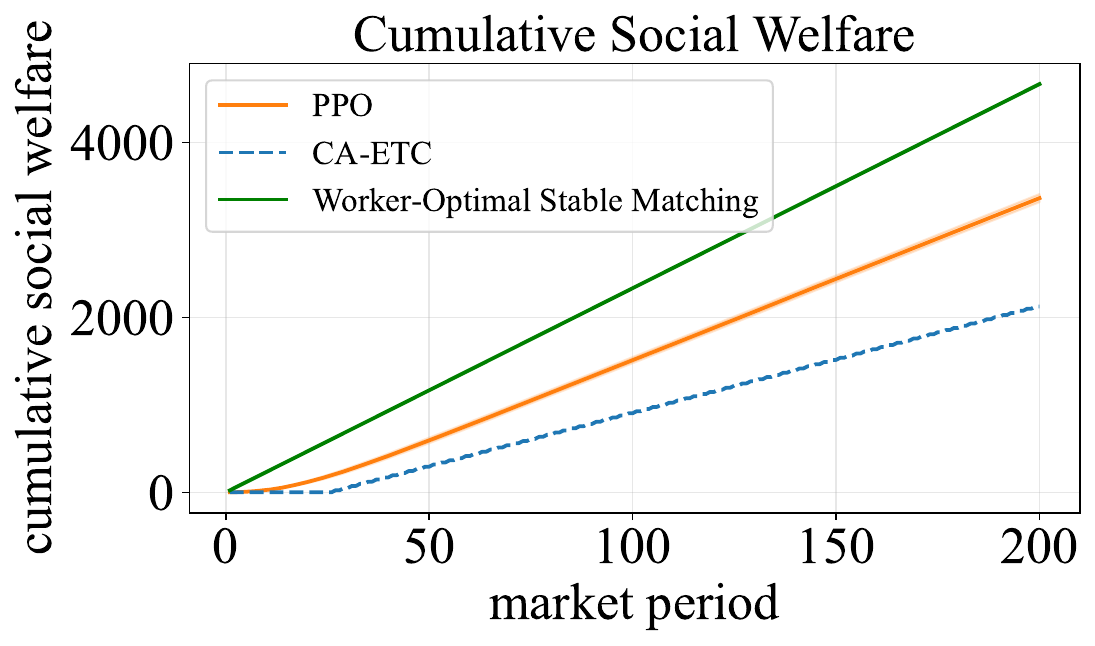}
        \caption{Social welfare}
        \label{fig:social_welfareno_noise}
    \end{subfigure}
    \hfill
    \begin{subfigure}{0.24\linewidth}
        \centering
        \includegraphics[width=\linewidth]{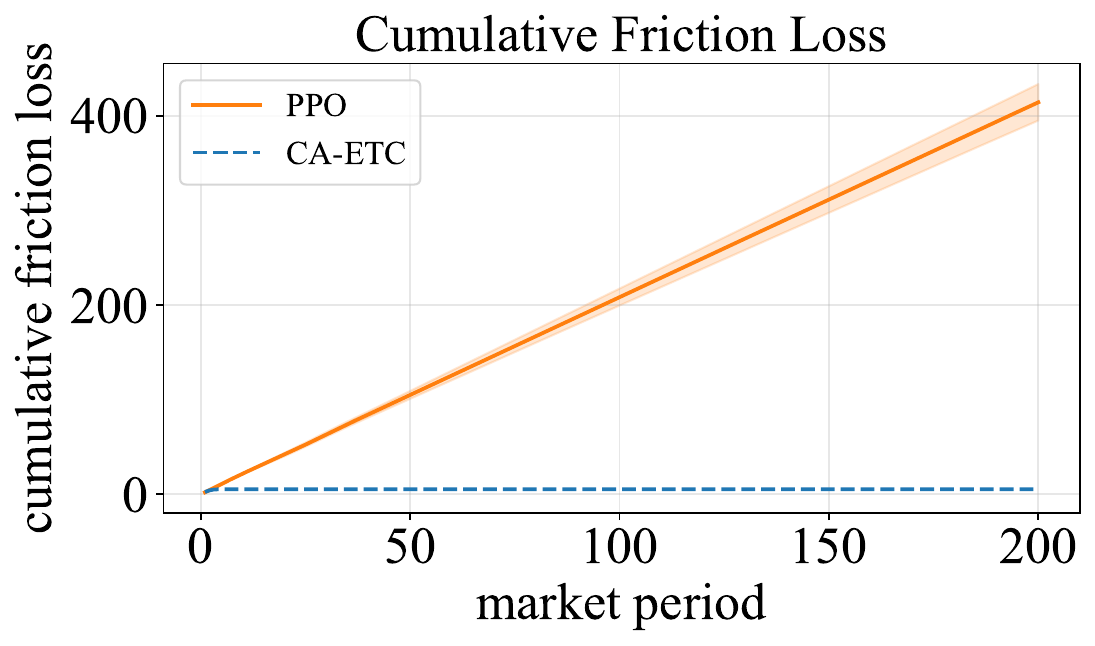}
        \caption{Friction loss}
        \label{fig:friction_lossno_noise}
    \end{subfigure}
    \caption{Comparison of \textsc{Learn2Match} (PPO) against CA-ETC in Low-noise / near-static setting in the small market. CA-ETC has near-zero cumulative friction loss. However, PPO still outperforms CA-ETC in both regret and social welfare.}
\label{fig:learn2match-vs-caetc-low-noise}
\end{figure}

\paragraph{Low-noise / near-static setting.}
Even in this favorable regime for CA-ETC, PPO achieves lower realized regret and higher realized welfare over the finite horizon (Figure~\ref{fig:learn2match-vs-caetc-low-noise}). At $t=200$, cumulative firm and worker regret are roughly halved relative to CA-ETC, showing the advantage of MARL algorithms on long-term planning. CA-ETC achieves near-zero cumulative friction loss, while PPO accumulates about ${\sim}400$. This is consistent with the design of CA-ETC: when feedback is nearly immediate and noiseless, explicit per-pair empirical averages and repeated Gale--Shapley recomputation recover a belief-induced stable matching that closely matches the latent stable benchmark. 

\begin{figure}[htbp]
    \centering
    \begin{subfigure}{0.24\linewidth}
        \centering
        \includegraphics[width=\linewidth]{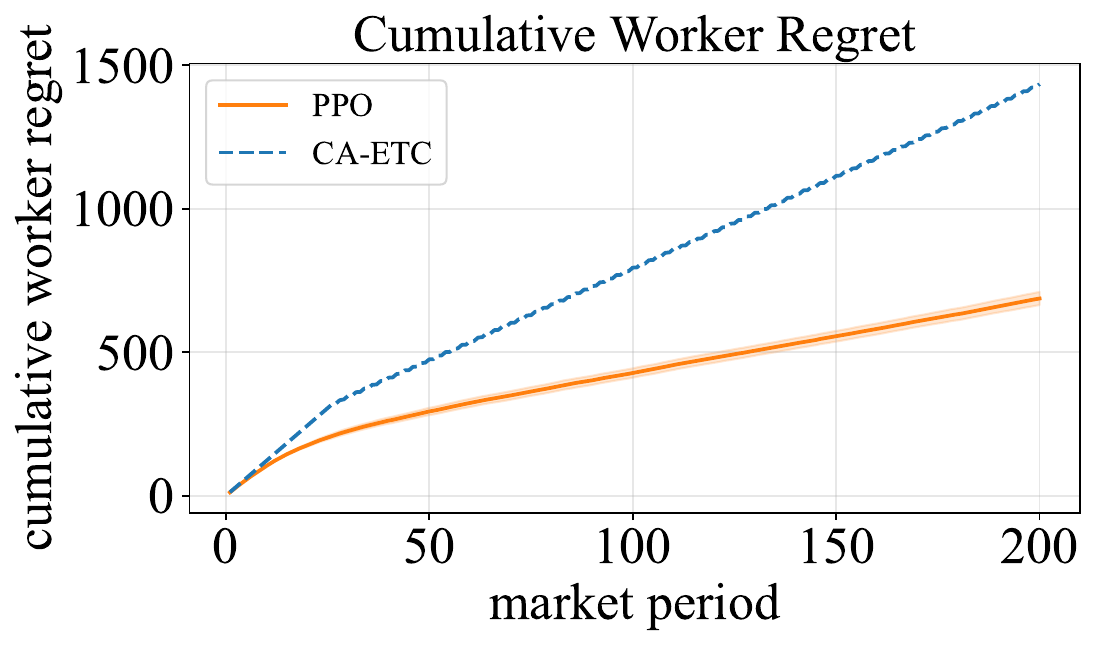}
        \caption{Worker regret}
        \label{fig:firm_regretfull_noise_graph}
    \end{subfigure}
    \hfill
    \begin{subfigure}{0.24\linewidth}
        \centering
        \includegraphics[width=\linewidth]{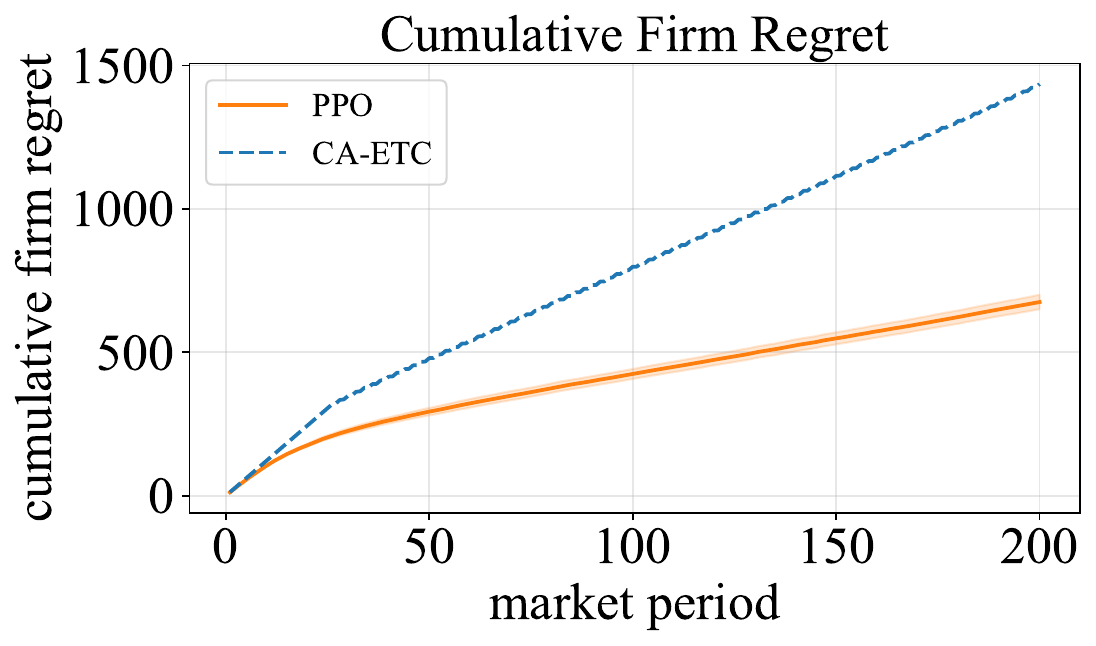}
        \caption{Firm regret}
        \label{fig:worker_regretfull_noise_graph}
    \end{subfigure}
    \hfill
    \begin{subfigure}{0.24\linewidth}
        \centering
        \includegraphics[width=\linewidth]{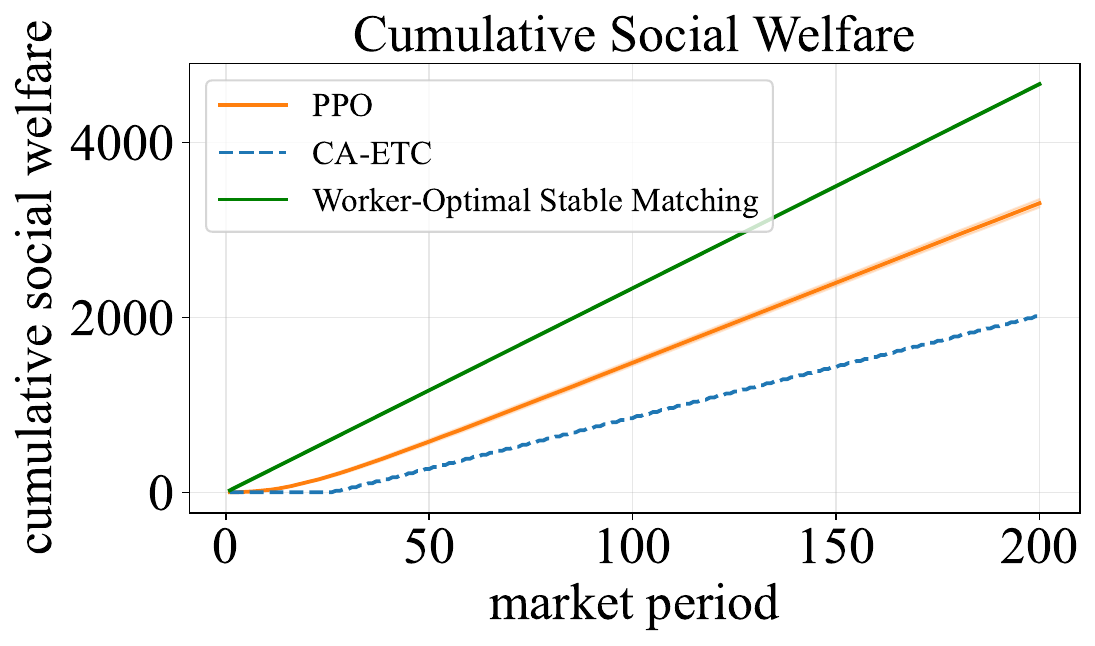}
        \caption{Social welfare}
        \label{fig:social_welfarefull_noise_graph}
    \end{subfigure}
    \hfill
    \begin{subfigure}{0.24\linewidth}
        \centering
        \includegraphics[width=\linewidth]{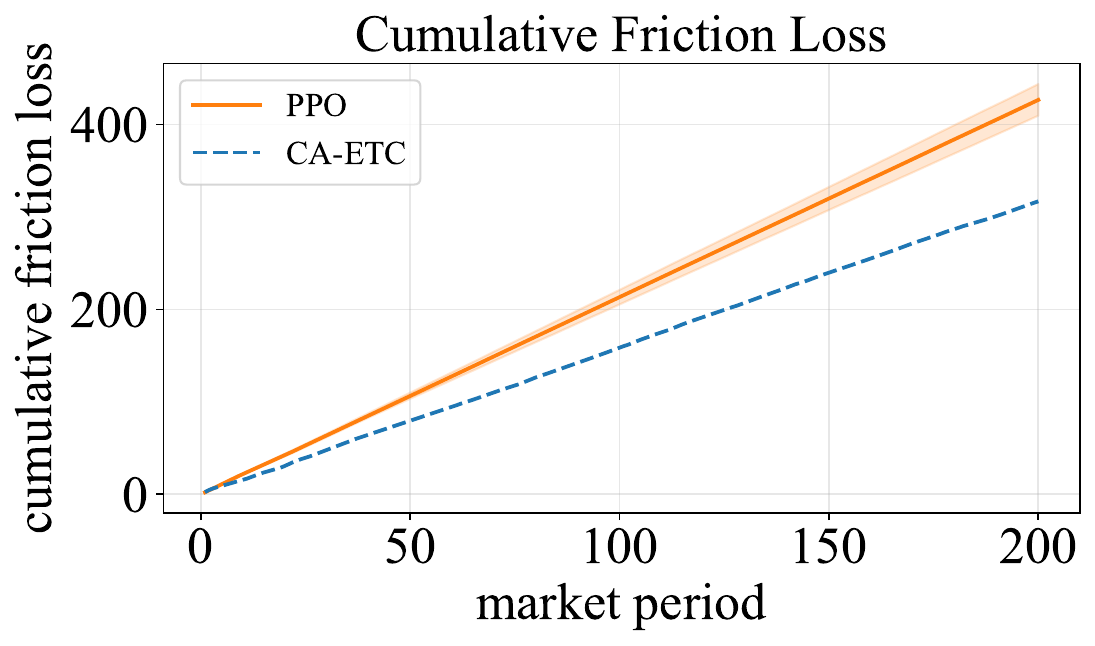}
        \caption{Friction loss}
        \label{fig:friction_lossfull_noise_graph}
    \end{subfigure}
    \caption{Comparison of \textsc{Learn2Match} (PPO) against CA-ETC in the temporally extended feedback setting in the small market. PPO outperforms CA-ETC in both regret and social welfare, but CA-ETC has lower friction loss.}
\label{fig:eval-rl-small}
\end{figure}

\begin{figure}[htbp]
    \centering
    \begin{subfigure}{0.24\linewidth}
        \centering
        \includegraphics[width=\linewidth]{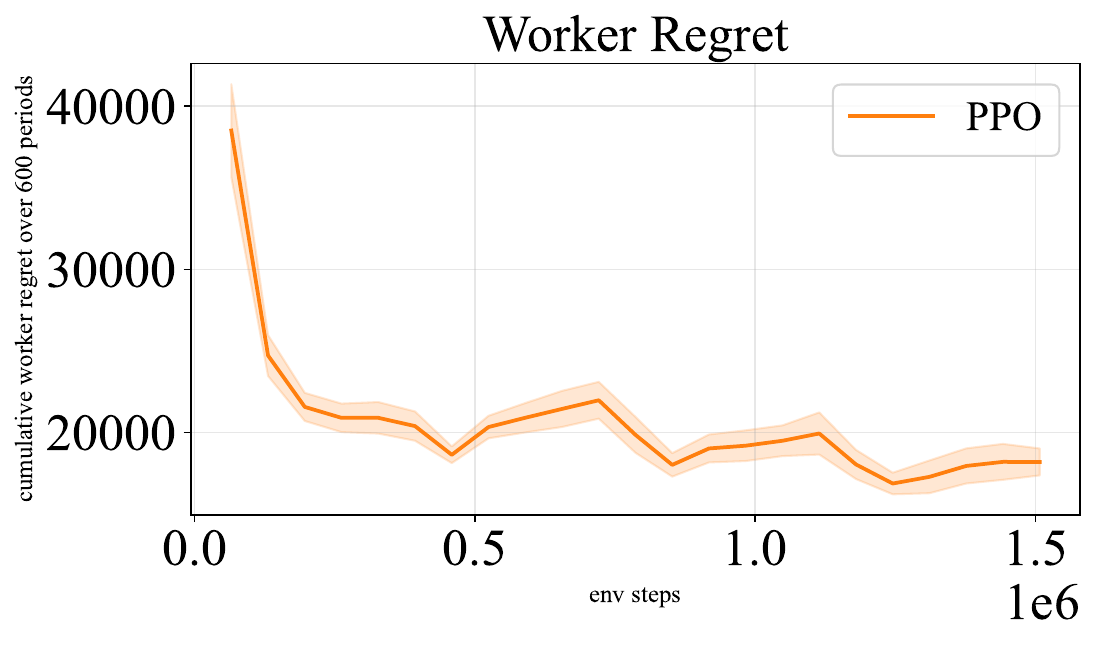}
        \caption{Worker regret}
        \label{fig:firm_regretlearning_curve}
    \end{subfigure}
    \hfill
    \begin{subfigure}{0.24\linewidth}
        \centering
        \includegraphics[width=\linewidth]{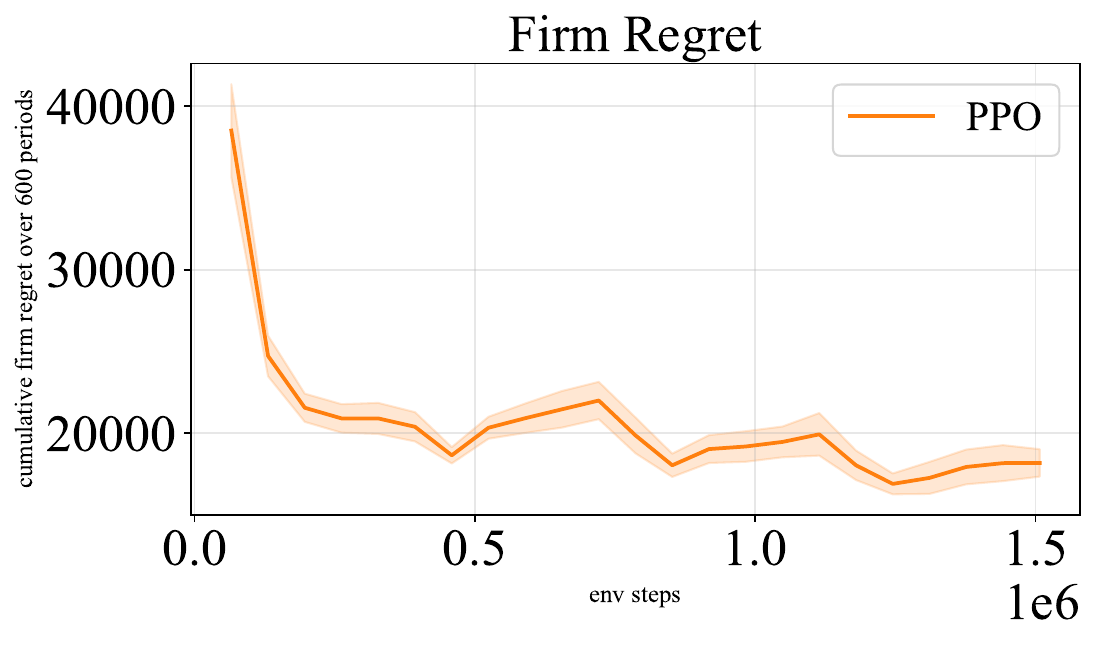}
        \caption{Firm regret}
        \label{fig:worker_regretlearning_curve}
    \end{subfigure}
    \hfill
    \begin{subfigure}{0.24\linewidth}
        \centering
        \includegraphics[width=\linewidth]{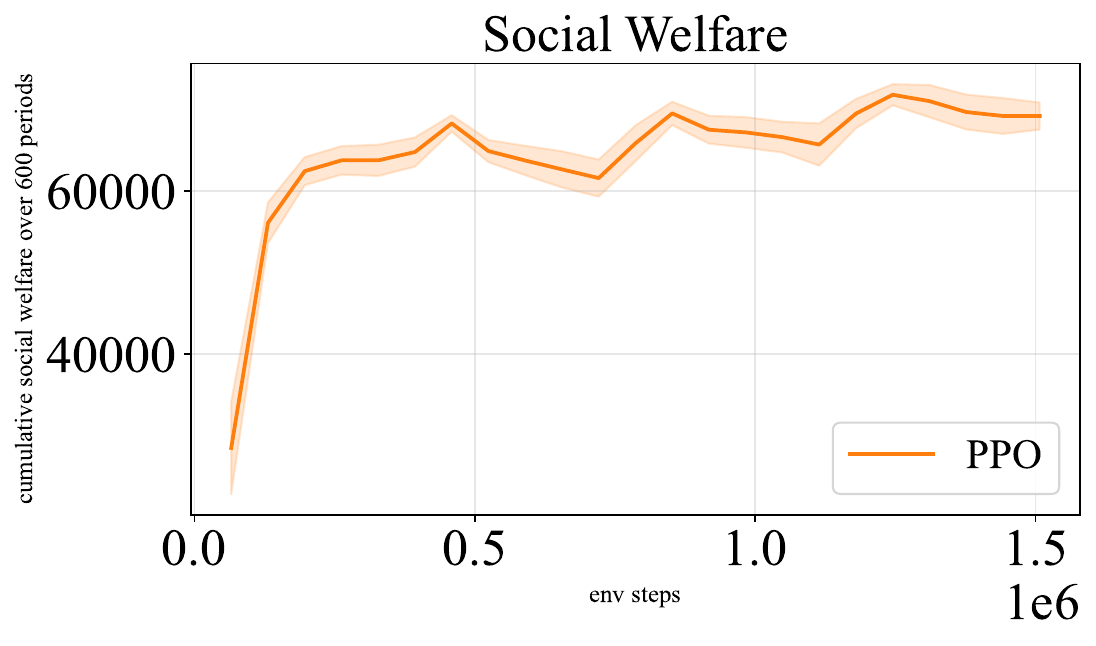}
        \caption{Social welfare}
        \label{fig:social_welfarelearning_curve}
    \end{subfigure}
    \hfill
    \begin{subfigure}{0.24\linewidth}
        \centering
        \includegraphics[width=\linewidth]{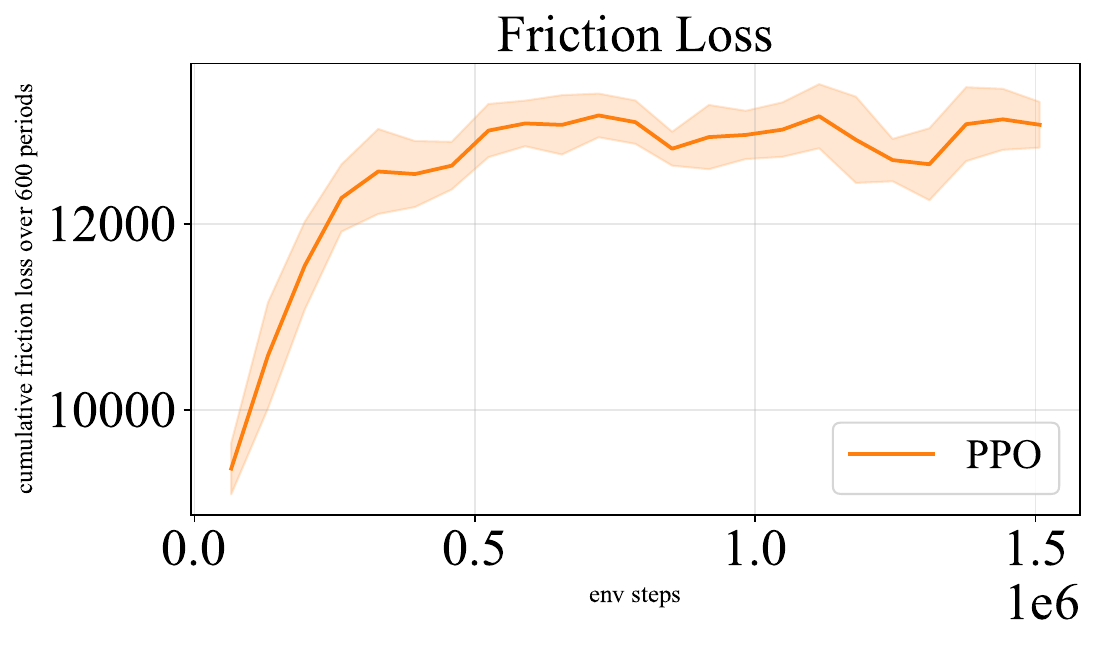}
        \caption{Friction loss}
        \label{fig:friction_losslearning_curve}
    \end{subfigure}
\caption{PPO learning curves in the large market, temporally extended feedback setting. Worker regret and firm regret decrease over training; social welfare increases and then converges. Friction loss converges to a non-zero value.}
\label{fig:learning-curve-large}
\end{figure}

\begin{figure}[htbp]
    \centering
    \begin{subfigure}{0.24\linewidth}
        \centering
        \includegraphics[width=\linewidth]{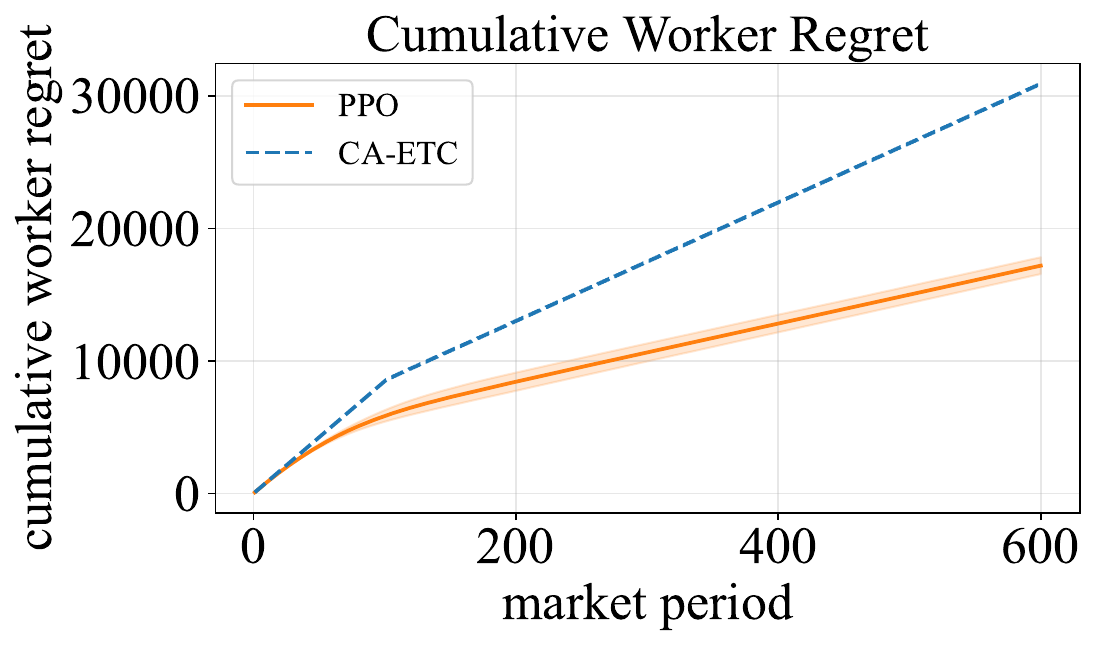}
        \caption{Worker regret}
        \label{fig:firm_regretlarge_market}
    \end{subfigure}
    \hfill
    \begin{subfigure}{0.24\linewidth}
        \centering
        \includegraphics[width=\linewidth]{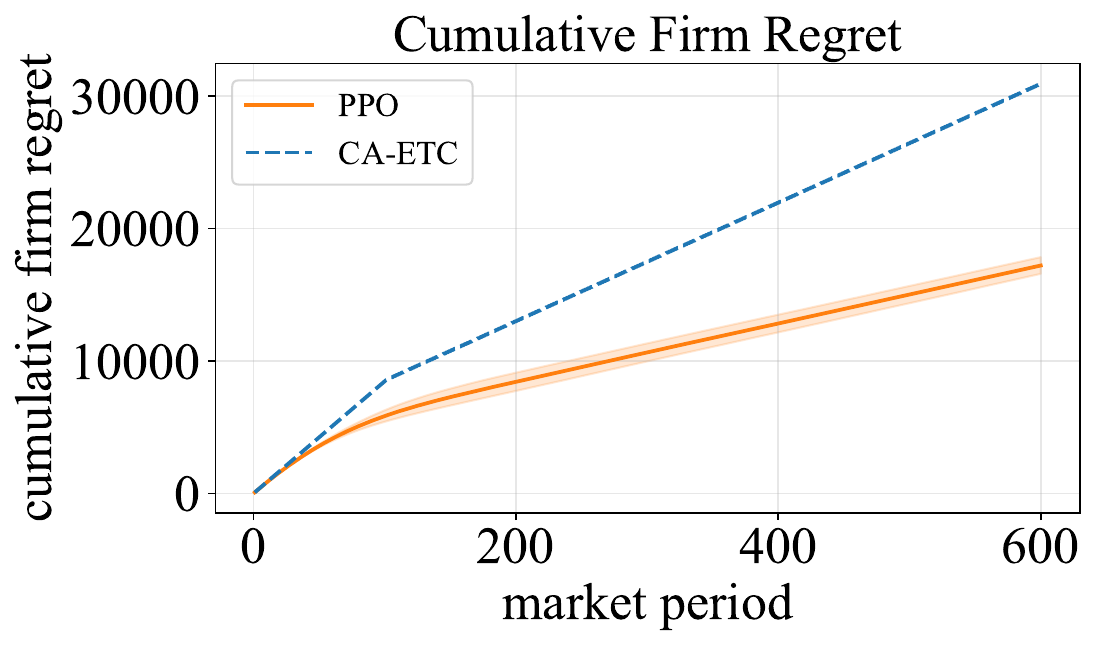}
        \caption{Firm regret}
        \label{fig:worker_regretlarge_market}
    \end{subfigure}
    \hfill
    \begin{subfigure}{0.24\linewidth}
        \centering
        \includegraphics[width=\linewidth]{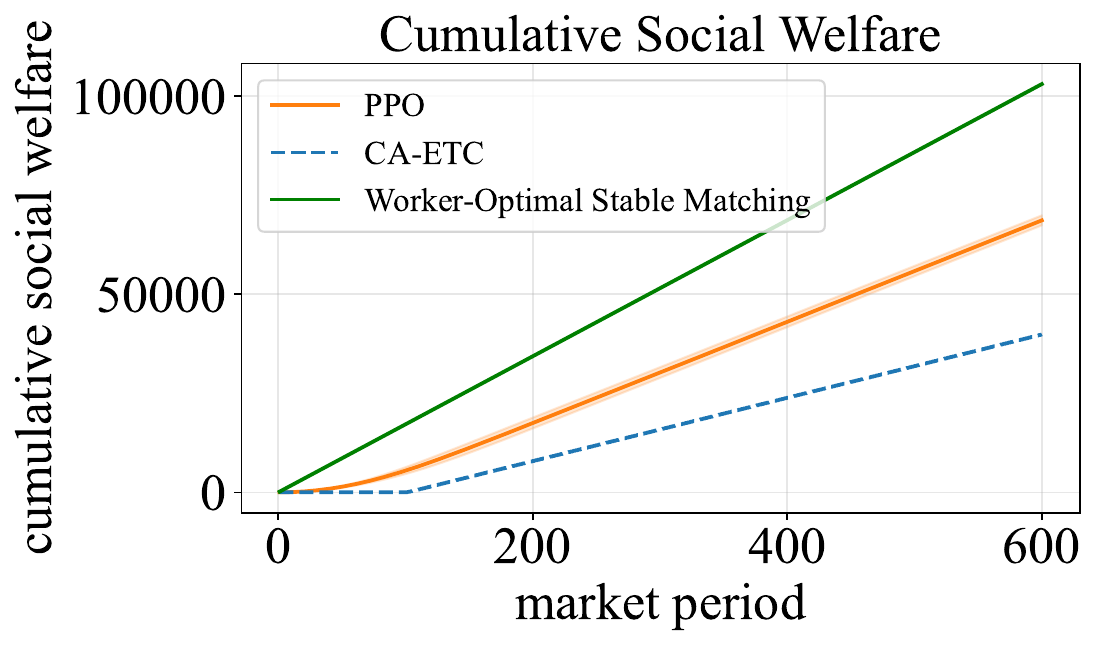}
        \caption{Social welfare}
        \label{fig:social_welfarelarge_market}
    \end{subfigure}
    \hfill
    \begin{subfigure}{0.24\linewidth}
        \centering
        \includegraphics[width=\linewidth]{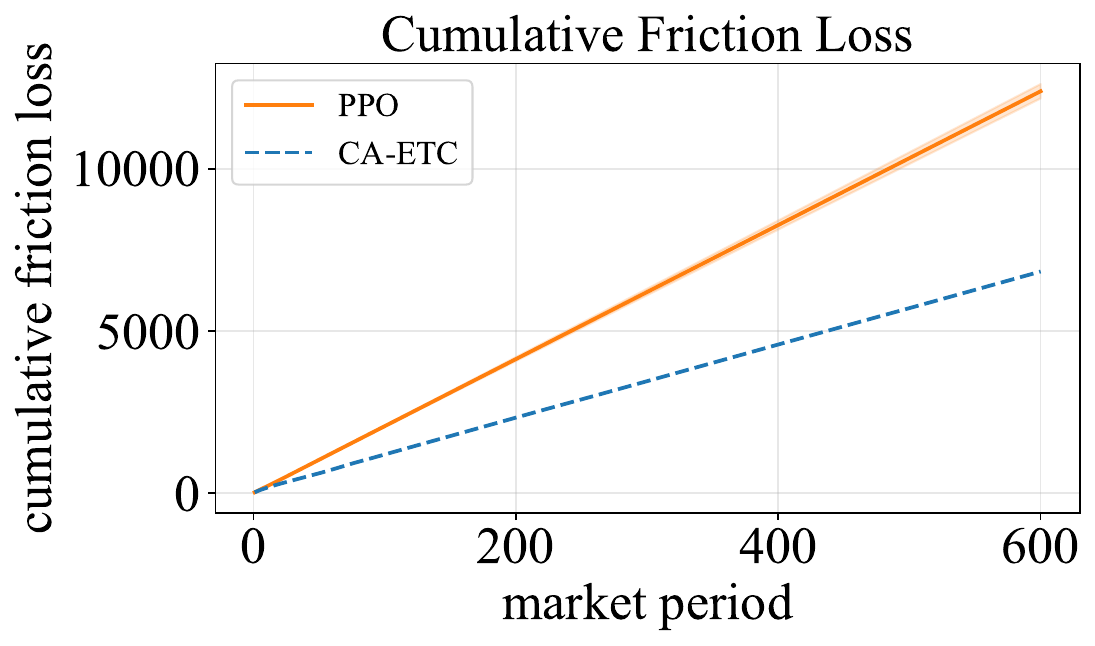}
        \caption{Friction loss}
        \label{fig:friction_losslarge_market}
    \end{subfigure}
    \caption{Comparison of \textsc{Learn2Match} (PPO) against CA-ETC in the temporally extended feedback setting in the large market. The result is consistent with the small market. PPO outperforms CA-ETC in both regret and social welfare, but CA-ETC has lower friction loss.}
\label{fig:learn2match-large-market}
\end{figure}

\paragraph{Temporally extended feedback.} The main benchmark setting restores the structure motivated in the introduction: interviews are noisy, post-match observations are noisy, and latent profiles are revealed only gradually through tenure. In this regime, PPO again outperforms CA-ETC on three of the four evaluation metrics, both in the small market (Figure~\ref{fig:eval-rl-small}) and the large market (Figure~\ref{fig:learn2match-large-market}). 

This is the regime where temporally extended feedback matters most. CA-ETC estimates preferences from noisy, short-run observations and then executes Gale--Shapley-style exploitation from those estimates. The welfare and regret gains therefore support the paper's central claim that matching markets with delayed revelation require algorithms that act while information is still unfolding, rather than algorithms that reduce the problem to one-shot preference estimation.

\begin{figure}
    \centering
    \includegraphics[width=1\linewidth]{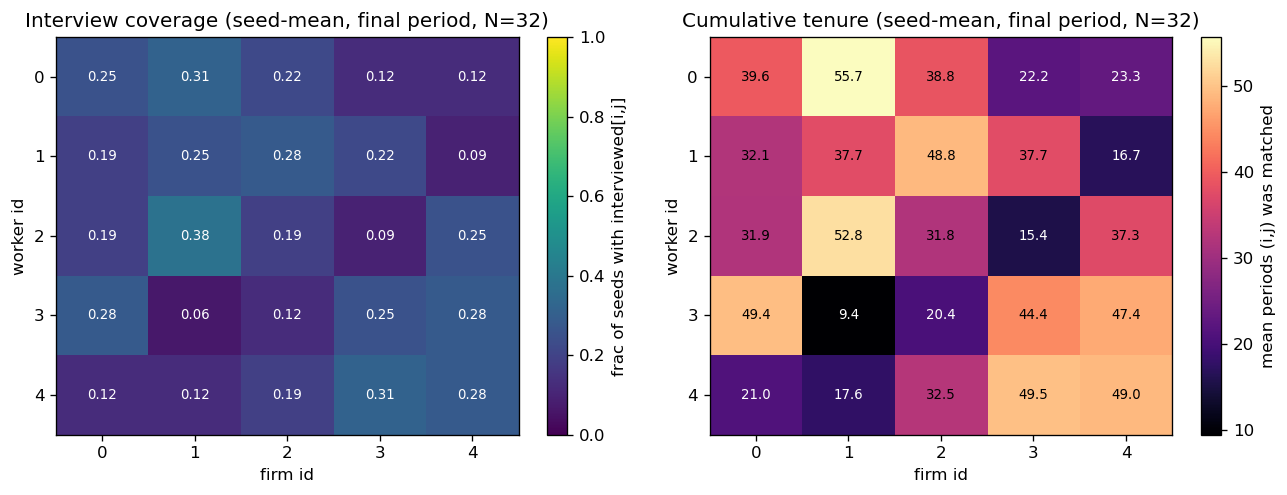}
        \caption{%
        \textit{Left:} interview coverage---the fraction that each pair $(i, j)$ was interviewed at least once by the end of the episode across all evaluation environments.
        \textit{Right:} mean cumulative tenure of each pair at the final period.
        Both figures are from the large market setting.
    }
    \label{fig:small_heatmap}
\end{figure}

Notably, although PPO achieves higher performance, the friction loss does not converge to zero, indicating insufficient exploration. Figure~\ref{fig:small_heatmap} explains this gap: PPO interviews a few pairs and then commits for long periods to promising matches, echoing the \emph{career path dependence }documented in labor economics, where relationships become more persistent with tenure and current matches shape future learning and mobility costs~\citep{farber1996learning,topel1992job,miller1984job,kambourov2009occupational}. This improves realized welfare through tenure-based learning, but leaves many unobserved pairs near their priors, highlighting the need to combine PPO-like adaptivity with CA-ETC-like coordinated exploration and stable-matching structure.

\section{Conclusion}
We introduced \textsc{Learn2Match}, a benchmark for studying two-sided matching markets with temporally extended feedback. Unlike standard learning-based matching models that treat proposals or matches as producing immediate noisy feedback about fixed preferences, our formulation models matching as a partially observable Markov game in which information is revealed gradually through sequential interaction. Our experiments show that reinforcement-learning policies better exploit temporal structure than bandit-style baselines by allocating interviews toward informative future matches, rematching early under uncertainty, and reducing information-friction loss over time.

\clearpage
\newpage

\bibliographystyle{plainnat}
\bibliography{reference}

\newpage
\appendix
\section{Implementation details}

\subsection{PPO implementation details}

Both small and large markets use an outside-option penalty of $-10^9$. Latent profiles $x_i$ and $y_j$ are drawn i.i.d. from $\mathcal N(0,I_d)$ and truncated to be non-negative.

The PPO implementation follows the following hyperparameters:

\begin{table}[H]
    \centering
    \caption{PPO hyperparameters used for all three multi-seed runs.}
    \label{tab:ppo-hyperparams}
    \begin{tabular}{lr}
        \toprule
        Hyperparameter & Value \\
        \midrule
        batch size & 4096 \\
        chunk length & 16 \\
        Recurrent layer type & GRU \\
        number of Recurrent layers & 1 \\
        GAE $\lambda$ & 0.95 \\
        discounting factor $\gamma$ & 0.997 \\
        hidden size & 64 \\
        learning rate & $3 \times 10^{-4}$ \\
        parallel environments & 128 \\
        ppo epochs & 4 \\
        ratio clip & 0.2 \\
        \bottomrule
    \end{tabular}
\end{table}

\subsection{For CA-ETC}

\begin{table}[H]
    \centering
    \caption{CA-ETC baseline hyperparameters used for all runs.}
    \label{tab:ca-etc-hyperparams}
    \begin{tabular}{lc}
        \toprule
        Hyperparameter & Value \\
        \midrule
        exploration block length $T_0$ & $\lceil N^{\mathcal{W}} / N^{\mathcal{F}} \rceil \cdot N^{\mathcal{F}}$ \\
        exploration schedule rate $\gamma$ & 0.4 \\
        confidence radius scale & 0.5 \\
        confidence check & enabled \\
        evaluation environments per run & 32 \\
        \bottomrule
    \end{tabular}
\end{table}

\paragraph{CA-ETC baseline in \textsc{Learn2Match}.}
We adapt the epoch-based CA-ETC algorithm to \textsc{Learn2Match} by treating agents on side $\mathcal W$ as players and agents on side $\mathcal F$ as arms. The algorithm proceeds in epochs, each consisting of an exploration block followed by an exploitation block. The exploration block collects empirical rewards for worker--firm pairs, and the exploitation block computes and executes a Gale--Shapley matching using the empirical preferences induced by these rewards.

For each pair $(i,j)\in [N^{\mathcal W}]\times [N^{\mathcal F}]$, the baseline maintains the empirical average rewards received by worker $i$ and firm $j$ when they are matched. These averages estimate
\[
    R_i^{\mathcal W}(s)=\langle x_i,\hat y_{ij}(s)\rangle,
    \qquad
    R_j^{\mathcal F}(s)=\langle \hat x_{ij}(s),y_j\rangle .
\]
They are updated only from realized reward feedback. The CA-ETC baseline records reward feedback at the moment a scheduled worker--firm pair forms a match. Specifically, once worker $i$ proposes to firm $j$ and firm $j$ accepts, the baseline records the match-induced reward signals
\[
    R_i^{\mathcal W}(s_t),\qquad R_j^{\mathcal F}(s_t),
\]
and updates the empirical averages for pair $(i,j)$. Whether the pair later remains matched through the dissolution phase affects the realized payoff in \textsc{Learn2Match}, but it does not affect whether CA-ETC has obtained a reward sample for preference estimation.

During the exploration block of epoch $\ell$, workers select firms according
to a collision-avoidance round-robin schedule, which under
$T_{0} \geq \lceil N_{w}/N_{f} \rceil \cdot N_{f}$ visits every worker--firm
pair $(i,j)$ at least once per block. Each scheduled pull is embedded into
one \textsc{Learn2Match} market period: worker $i$ conducts the scheduled
interview with firm $j$, drawing the noisy belief signals
$\hat{y}_{i,j} = y_{j} + \varepsilon^{(f)}_{i,j}$ and
$\hat{x}_{i,j} = x_{i} + \varepsilon^{(w)}_{i,j}$ with
$\varepsilon^{(\cdot)}_{i,j} \sim \mathcal{N}(0,\sigma_{\mathrm{interview}}^{2} I_{d})$;
worker $i$ then proposes a match and firm $j$ accepts, forming the tentative
pair. The belief-induced utilities
$\langle x_{i}, \hat{y}_{i,j} \rangle$ and
$\langle \hat{x}_{i,j}, y_{j} \rangle$ are recorded as the CA-ETC sample for
$(i,j)$ and incorporated into the running empirical means. The tentative pair is released at the subsequent retention step, as the
exploration block aims only to measure
$\langle x_{i}, \hat{y}_{i,j} \rangle$ and $\langle \hat{x}_{i,j}, y_{j} \rangle$
for each scheduled pair; releasing the agents frees them for the next
round-robin pull.

After the exploration block, every agent holds an empirical estimate of its
preferences over the opposite side, which is passed to the Gale--Shapley
routine that begins the exploitation block.

During the exploitation block, the baseline runs worker-proposing Gale--Shapley with the current empirical preference lists. The resulting matching is not imposed directly on the environment state; it is executed through the \textsc{Learn2Match} action protocol. Each worker proposes to its Gale--Shapley assigned firm, each firm accepts its assigned worker and rejects other proposals, and matched pairs do not voluntarily dissolve during the exploitation block. The exploitation block does not perform interviews or collect additional preference information. At the end of the exploitation block, the next epoch begins with a new exploration block, after which empirical preferences and the Gale--Shapley matching are recomputed.

Thus, the adaptation preserves the CA-ETC structure
\[
\underbrace{\text{interview} + \text{tentative match}}_{\text{Exploration (epoch } \ell\text{)}}
\;\longrightarrow\;
\underbrace{\text{LCB/UCB ranking}}_{\text{Preference estimation}}
\;\longrightarrow\;
\underbrace{\text{GS matching} + \text{retention}}_{\text{Exploitation (epoch } \ell\text{)}}
\;\xrightarrow{\ell \mapsto \ell+1}\; \cdots
\]
repeated across epochs. The only substantive change is that each CA-ETC exploration pull of pair $(i,j)$ must be received through the valid \textsc{Learn2Match} interaction protocol: interview, match proposal, match acceptance but without retention, and reward recording. Exploitation only computes Gale--Shapley from the resulting empirical preference lists and executes the selected matching through standard \textsc{Learn2Match} match proposals and acceptances.

\begin{figure}[H]
    \centering
    \includegraphics[width=0.5\linewidth]{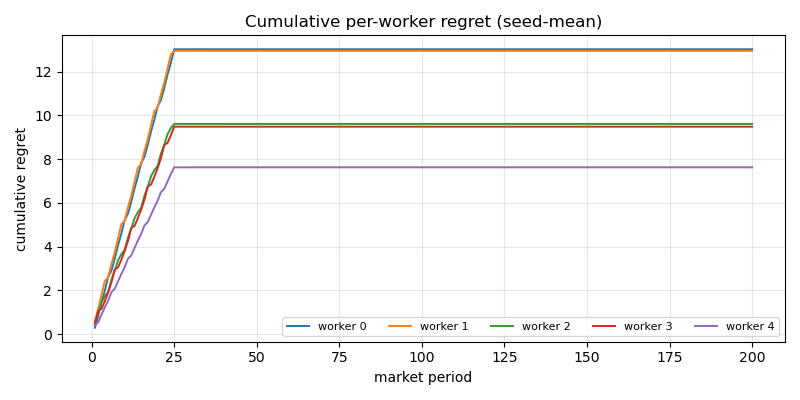}
    \caption{Cumulative per-worker regret of CA-ETC inside \textsc{Learn2Match} under the low-noise setting. Each line is one worker, seed-mean across runs. Regret plateaus once the algorithm commits to its empirical Gale--Shapley matching}
    \label{fig:caetc-worker}
\end{figure}

\begin{figure}[H]
    \centering
    \includegraphics[width=0.5\linewidth]{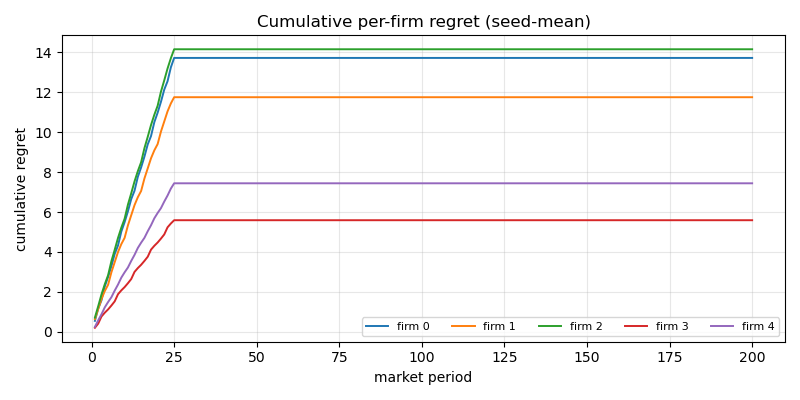}
    \caption{Cumulative per-firm regret of CA-ETC inside \textsc{Learn2Match} under the low-noise setting, with the same plateau behavior as Figure~\ref{fig:caetc-worker}.}
    \label{fig:caetc-firm}
\end{figure}

\paragraph{CA-ETC under the low-noise setting.}
We run CA-ETC inside \textsc{Learn2Match} under \texttt{--Nw 5 --Nf 5 --d 3 --horizon 200 --sigma\_interview 0.001 --sigma\_match 0 --lambda\_reveal 100}. Figures~\ref{fig:caetc-worker} and~\ref{fig:caetc-firm} show that cumulative per-worker and per-firm regret rise during the initial exploration block and then flatten as the algorithm commits to its empirical Gale--Shapley matching, reproducing the structure reported in~\citep{pagare2024explore}. We prove that we are able to show that we can cover exactly the same CA-ETC results under some parameter setting.

\begin{figure}
    \centering
    \includegraphics[width=1\linewidth]{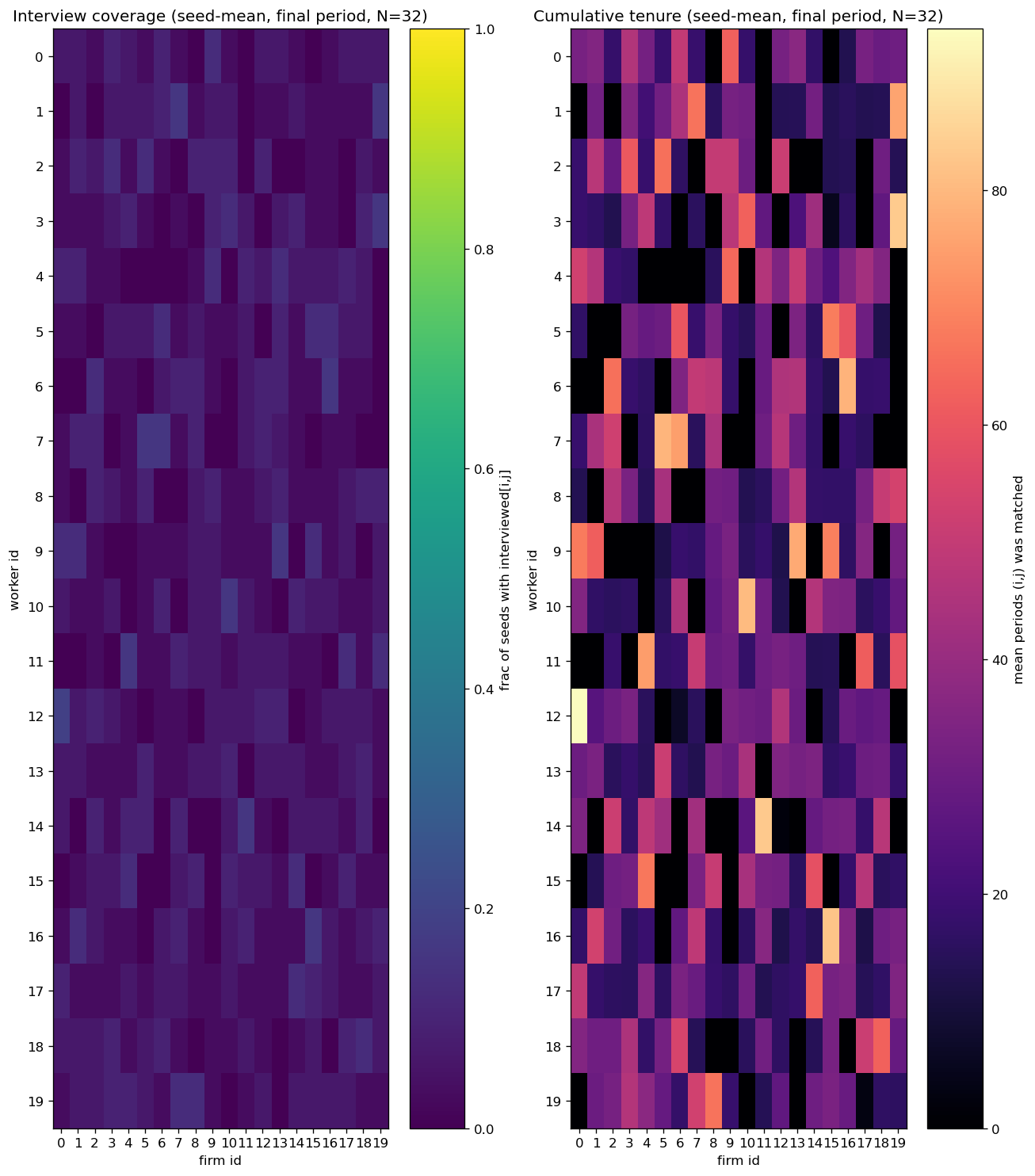}
    \caption{\textit{Left:} interview coverage. \textit{Right:} mean cumulative tenure of each pair at the final period.}
    \label{fig:heatmap}
\end{figure}

\section{Broader Impact and Limitations}

L\textsc{earn2}M\textsc{atch} provides a shared testbed for the matching and MARL communities to develop algorithms closer to the structure of real markets such as labor, residency, school choice, dating, and ride-hailing, with the potential to reduce search frictions and information-friction loss for participants on both sides. As with any matching system operating in high-stakes domains, learned policies may inherit biases in observed preferences and create incentives for strategic behavior, which is why we view the benchmark as a controlled research environment rather than a deployable system. Compared to real-world matching markets with millions of users, our experiments are limited to small markets with one-to-one matching, leaving the scaling behavior of PPO and CA-ETC in larger and many-to-many markets open. The benchmark also relies on specific functional-form choices---Gaussian latent profiles, linear rewards, and sigmoidal revelation---that capture temporally extended feedback tractably but do not reflect the non-Gaussian preferences and heterogeneous learning rates of real markets. Finally, we evaluate only one representative MARL method against one bandit baseline on synthetic data, and view broader algorithmic comparisons, alternative interaction protocols, and validation on real matching data as important directions for future work.

\clearpage
\newpage

\end{document}